%% file: polis.tex
\documentclass[10pt,twocolumn,letterpaper]{article}

\usepackage[pagenumbers]{cvpr}   

\makeatletter
\def\@maketitle{%
  \newpage
  \null
  \vskip 0.18in
  \hrule height 1pt
  \vskip 0.16in
  \begin{center}
    {\LARGE\bf \@title \par}
  \end{center}
  \vskip 0.16in
  \hrule height 0.5pt
  \vskip 0.22in
  \begin{center}
    {\large \@author \par}
  \end{center}
  \vskip 0.6em
}
\def\abstract{%
  \centerline{\large\bf Abstract}%
  \vspace*{8pt}%
}
\makeatother

\usepackage[T1]{fontenc}
\usepackage{microtype}
\usepackage{amsfonts}
\usepackage{multirow}
\usepackage{stfloats}
\usepackage{cuted}
\usepackage{placeins}
\usepackage{float}
\usepackage[dvipsnames]{xcolor}
\makeatletter
\def\cvprsection{\@startsection {section}{1}{\z@}
   {-8pt plus -1pt minus -2pt}{5pt} {\large\bf}}
\def\cvprsubsection{\@startsection {subsection}{2}{\z@}
   {-6pt plus -1pt minus -2pt}{4pt} {\elvbf}}
\makeatother

\definecolor{cvprblue}{rgb}{0.21,0.49,0.74}
\usepackage{xurl}
\usepackage[pagebackref,breaklinks,colorlinks,citecolor=cvprblue]{hyperref}
\AtEndPreamble{%
  \crefname{appendix}{Appendix}{Appendices}%
  \Crefname{appendix}{Appendix}{Appendices}%
}

\def\paperID{*****}
\def\confName{3DV\xspace}
\def\confYear{2027\xspace}

\newcommand{\method}{Polis}

\title{\method{}: 3D Self-Supervision at City Scale}

\author{%
\textbf{Alexander M. Rusnak}\hskip 1.85em
\textbf{Sophia Kovalenko}\hskip 1.85em
\textbf{Jingru Wang}\\[0.7em]
\textbf{Ismail Moudden}\hskip 1.85em
\textbf{Xiru Wang}\hskip 1.85em
\textbf{Fr\'{e}d\'{e}ric Kaplan}\\[0.85em]
{\normalsize \'Ecole Polytechnique F\'ed\'erale de Lausanne}%
}

\begin{document}
\maketitle

\begin{abstract}
Reliable semantic representations derived from city-scale 3D models are increasingly important for urban analysis, infrastructure monitoring, autonomous systems, and heritage conservation. However, urban scenes of large spatial extent captured through aerial surveying differ substantially from the indoor, object-level, and self-driving LiDAR data used to pretrain most 3D self-supervised models. We introduce \method{}, to our knowledge the first application of Sketched Isotropic Gaussian Regularization (SIGReg) as an objective for a native point cloud encoder, and evaluate it through a frozen-feature benchmark spanning fourteen city- and building-scale corpora. \method{} combines geometrically matched cosine invariance, SIGReg~\cite{balestriero2025lejepa}, and VICReg-style anti-collapse terms~\cite{bardes2022vicreg} with a $12.8\mathrm{k}$-scene outdoor pretraining mixture and gravity-preserving spatial view sampling. Controlled ablations show that this objective outperforms student--teacher architecture alternatives, as well as Polis versions without anti-collapse terms, on the same representative outdoor corpus. On three pretraining-disjoint city datasets, \method{} reaches $23.8\%$ mean mIoU versus $16.3\%$ for the next-best encoder under high-capacity frozen probing, and $17.3\%$ versus $16.1\%$ at a matched point and voxel budget. The same city-scale lead holds on datasets whose training sets were seen in pretraining. On localized terrestrial captures with fine-grained facade and streetscape labels, the ranking reverses. Our results show that distributionally-regularized joint embedding architectures can be successful on challenging city-scale 3D scenes, and that transfer improves when self-supervision is designed for the capture geometry and spatial context of this domain while also revealing the limits of this specialization.
\end{abstract}

\begin{figure}[!t]
  \centering
  \includegraphics[width=0.92\linewidth]{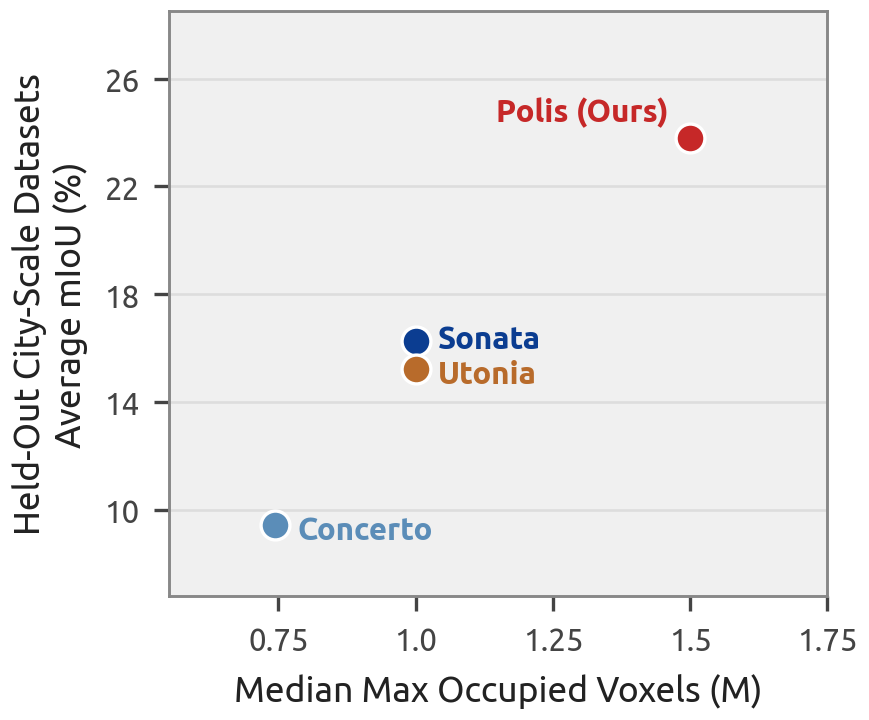}
  \caption{High-capacity frozen-probe mIoU on the three pretraining-disjoint city-scale datasets (UNS-Geo, HRHD-HK, and WHU-Urban3D) versus median whole-scene occupied-voxel capacity. Matched-budget scores and the remaining eleven corpora are in \cref{sec:main-results}.
  \textbf{Polis combines the strongest held-out city transfer with the largest typical capacity.}}
  \label{fig:capacity-transfer}
\end{figure}

\vspace{-12pt}
\section{Introduction}
\label{sec:intro}

Outdoor point clouds are derived from a diverse array of capture techniques and used for a variety of applications including urban planning, environmental monitoring, autonomous vehicles, infrastructure inspection, and heritage conservation. The scope of downstream usages makes city-scale point clouds an important target for representation learning with high transfer quality, yet their heterogeneous sensing, density, and spatial extent make them a difficult target for general-purpose 3D self-supervision. Thus far, the strongest 3D self-supervised recipes are trained mainly on indoor or object-level data, and, when they do utilize outdoor scenes, primarily focus on localized, egocentric self-driving LiDAR sequences. Within this targeted data scope, these stronger models have established a relatively consistent training paradigm based on self-distillation and, sometimes, bootstrapping their representations with frozen image encoders~\cite{wu2025sonata,zhang2025concerto,zhang2026utonia}.
Despite their broad ambitions, these models have not been systematically compared across many city-scale datasets or at the point and voxel budgets required by large outdoor scenes.
This leaves two coupled questions: how well do current frozen 3D SSL features survive the shift to city scale, and can a system specialized to outdoor data and geometry transfer more effectively in that regime?
We find that an outdoor-specialized, encoder-only SIGReg point encoder, trained with gravity-preserving views on a $12.8\mathrm{k}$-scene city mixture, yields stronger frozen transfer on broad aerial city scenes than current general-purpose 3D SSL systems, and weaker transfer on localized terrestrial part taxonomies.

In order to test this, we assembled a suite of city- and building-scale datasets. We utilize the nomenclature \emph{city-scale} for broad, predominantly aerial urban scenes and \emph{building-scale} for localized terrestrial captures (facade, street, or square) with finer part taxonomies; the split is the capture regime, not whether the subject is urban.

\begin{figure*}[t]
  \vspace*{-8pt}
  \centering
  \includegraphics[width=\textwidth]{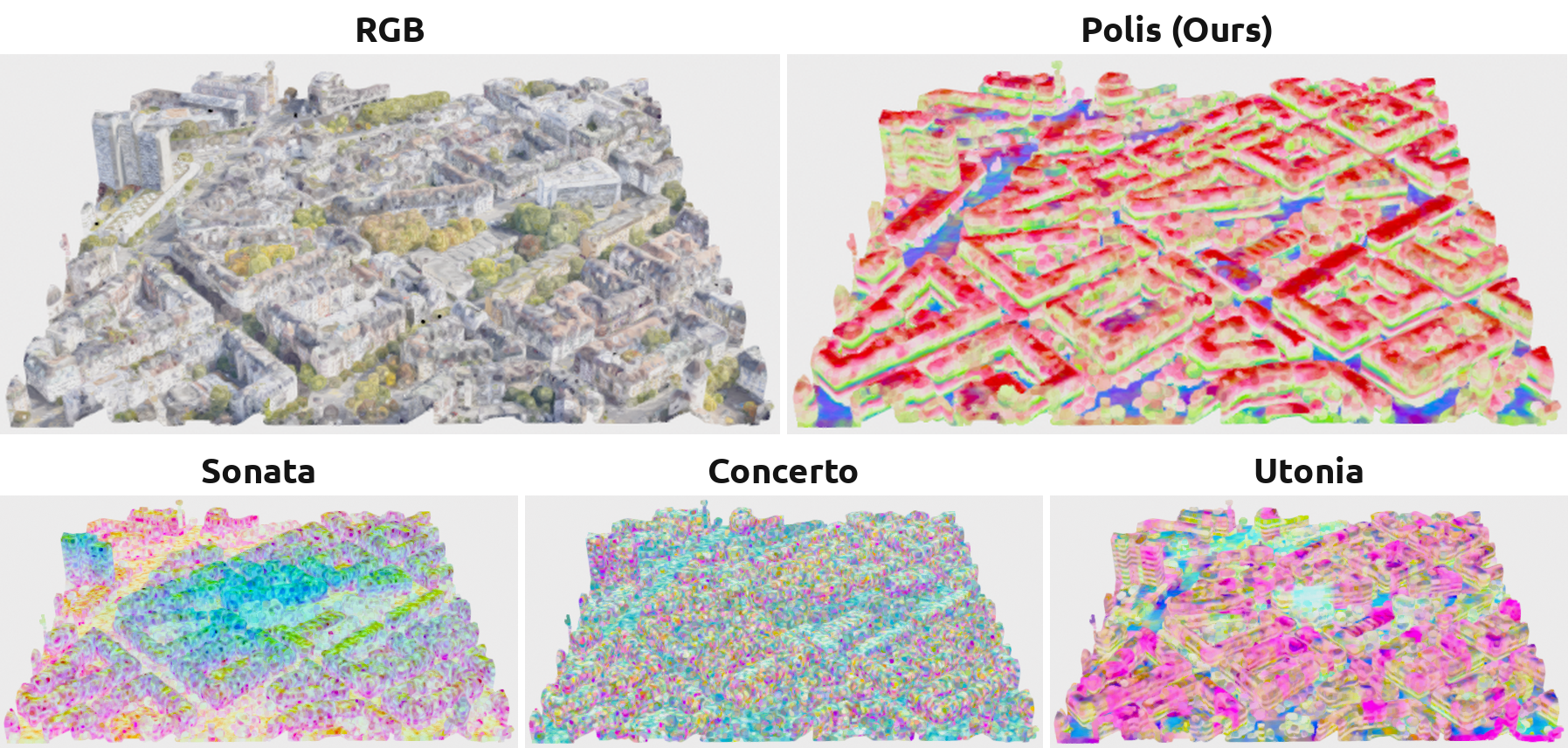}
  \caption{Frozen-backbone feature PCA on a Zurich scene from Swiss3DCities,
  shown from a matched oblique view.
  Top: RGB and \textbf{\method{} (ours)}. Bottom: Sonata, Concerto, and Utonia.
  All models use their high-budget point/voxel capacity.
  \textbf{Polis produces more semantically and spatially coherent outdoor feature structure in this example, while the baselines exhibit weaker separation or visible spatial artifacts.}}
  \label{fig:qual-zurich}
\end{figure*}

This paper addresses both questions through a complete pretraining system, \method{}, and a comprehensive evaluation protocol.
The name \method{} follows Aristotle's account of the \emph{polis} as the complete community whose whole is prior to its parts~\cite{aristotle1944politics}: households compose villages, and villages compose the city, but those smaller units become intelligible only through the larger order they jointly constitute.
We adopt that hierarchical similitude for city-scale representation learning.
Our encoder aggregates voxels into coarser tokens and then into scene-level structure, and our attentive probe classifies each local feature only after concatenating a global descriptor; and one of our central empirical findings is that strong city-scale transfer is facilitated by more expansive scene-level context rather than denser, local decoding.
\method{}'s training mixture spans outdoor and city-scale sources, balanced across corpora and sampled from persistent spatial chunks via per-pass area crops and partially gravity-preserving views.
Its native LitePT-L point encoder~\cite{yue2026litept} uses a single-encoder objective inspired by LeJEPA~\cite{balestriero2025lejepa}, combining SIGReg with cosine invariance between global and local tokens matched by nearest neighbors in their original, pre-augmentation coordinates, together with VICReg-style variance/covariance control~\cite{bardes2022vicreg}.
We then benchmark \method{}, Sonata, Concerto, and Utonia under a unified frozen-probe codepath, both at a shared point/voxel budget decoding across smaller spatial windows and at each model's largest practical one-shot scene budget.
Because these systems differ in data, backbone, modalities, preprocessing, and objective, the comparison measures complete-system transfer, while the ablations isolate whether the objective itself is appropriate for city-scale self-supervised learning.

\vspace{-10pt}
\paragraph{Contributions.}
\begin{enumerate}[leftmargin=*,itemsep=0.15em]
  \item A broad benchmarking of 3D SSL models across city- and building-scale datasets, with matched-budget and model-specific high-capacity protocols that expose both transfer quality and practical scene capacity.
  \item Evidence that an outdoor-specialized system transfers strongly to pretraining-disjoint, broad-extent city datasets while remaining weaker on localized terrestrial and facade-part tasks, together with analyses of potential radial or vertical biases and semantics.
  \item An outdoor-specialized pretraining system: a 22-corpus city mixture, gravity-preserving spatial view sampling, and a SIGReg-based, single-encoder objective for a native point encoder.
\end{enumerate}
Figure~\ref{fig:capacity-transfer} summarizes held-out city transfer:
\method{} has the highest mean mIoU and realizes the largest
median whole-scene occupied-voxel budget.

\section{Related Work}
\label{sec:related}

\begin{figure*}[t]
  \vspace*{-14pt}
  \centering
  \includegraphics[width=\textwidth]{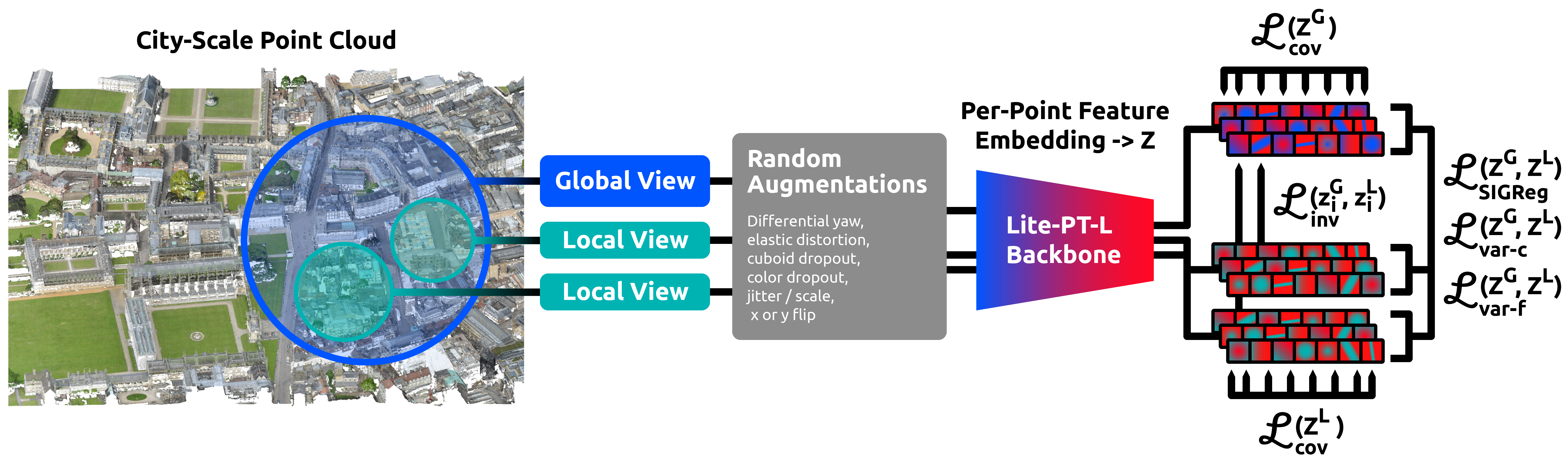}
  \caption{\method{} training stack.
  An area crop from a sampled outdoor scene (or a persistent spatial chunk of a large scene) yields one global and two local spatial views
  (origin coordinates retained for matching), which receive independent
  gravity-preserving geometric and chromatic augmentations
  (differential yaw/tilt, elastic distortion, cuboid and color dropout,
  jitter/scale, and flips) before a shared LitePT-L encoder.
  Training combines matched cosine invariance $\mathcal{L}_{\mathrm{inv}}$
  over origin-coordinate nearest-neighbor pairs with SIGReg and
  VICReg-style variance/covariance terms on the global and concatenated-local
  feature packs (\cref{eq:total-loss}).}
  \label{fig:architecture}
\end{figure*}

\paragraph{3D Self-Supervised Learning.}
PointContrast~\cite{xie2020pointcontrast}, Masked Scene Contrast~\cite{wu2023msc}, and Point-MAE~\cite{pang2022pointmae} established contrastive and masked objectives for points, with PointContrast already a joint-embedding recipe that pairs corresponding 3D points across views.
Joint-Embedding Predictive Architectures (JEPAs) learn relations between latent representations across views~\cite{lecun2022path,assran2023ijepa}, building on a longer line of work that trains networks to predict one view's latent code from another's~\cite{becker1992self,schmidhuber1992learning}.
Early 3D JEPA architectures, Point-JEPA~\cite{sait2025pointjepa} and 3D-JEPA~\cite{hu20243djepa}, use predictor-based point recipes but learn only on small object datasets, and they differ in where their targets originate: 3D-JEPA pairs a context-aware decoder with dVAE-token targets from a tuned 3D autoencoder, rather than the EMA target encoder of Point-JEPA.
More recently, DINO-style self-distillation~\cite{caron2021dino,oquab2024dinov2} has become the dominant paradigm for 3D SSL, with the strongest models coupling a student to a momentum or EMA teacher through stop-gradient updates and carefully tuned centering and temperature schedules, while scaling the recipe to large mixed corpora~\cite{wu2025sonata,zhang2025concerto,zhang2026utonia}; \cref{tab:pretrain-scale} gives their sizes and objectives.
However, self-distillation has recently been challenged by a new approach to pretraining for image representations, LeJEPA, which replaces that student--teacher stack with a distributional constraint~\cite{balestriero2025lejepa}: it argues that the isotropic Gaussian is the optimal embedding distribution for minimizing downstream prediction risk, and introduces sketched isotropic Gaussian regularization (SIGReg) to enforce it.
Although JEPA implementations often carry a latent predictor, the LeJEPA formulation does not require one: its Table~4 shows that the predictor and teacher--student architecture can be removed without collapse~\cite[Table~4]{balestriero2025lejepa}.
\method{} follows LeJEPA's direct cross-view alignment, though our ablations show a latent predictor does not strongly degrade transfer and slightly stabilizes training (\cref{sec:ablation-ema-pred}).
Point correspondence and the full objective are specified in \cref{sec:loss}.
Prior 3D SSL works report strong general-purpose representations, and Utonia trains a single encoder across domains that include outdoor and urban scenes, but we are not aware of a comparably broad cross-model evaluation focused on city-scale scenes and the practical capacity limits of released encoders.

\vspace{-4pt}
\paragraph{Backbones and probe architectures.}
Among our evaluated models, all but \method{} utilize a backbone from the vanilla Point Transformer family~\cite{zhao2021pointtransformer,wu2024ptv3}. \method{} uses a more recent variant of the Point Transformer, LitePT~\cite{yue2026litept}, which replaces attention in the initial encoder blocks with convolutions, moving the more expensive attention calculation deeper into the network where substantial hierarchical compression has already reduced the number of operational tokens; they also introduce a novel rotary positional encoding, PointROPE~\cite{yue2026litept,su2024rope}.

\begin{table}[H]
  \centering
  \caption{Pretraining scale, schedule, and whether a pretrained DINOv2 image tower is used for bootstrapping representation quality.
  Baseline asset and schedule figures are from the cited papers; \method{} figures are from the headline training run.}
  \label{tab:pretrain-scale}
  \scriptsize
  \setlength{\tabcolsep}{1.5pt}
  \begin{tabular}{@{}llccc@{}}
    \toprule
    Model & Assets & Ep.\ / bs / GPUs & DINOv2 & Domain \\
    \midrule
    Sonata~\cite{wu2025sonata}   & $\sim$140k scenes & 200 / 96 / 32 & no & indoor+sim \\
    Concerto~\cite{zhang2025concerto} & $\sim$40k clouds + 300k img & 100 / -- / 16 H20 & yes & indoor \\
    Utonia~\cite{zhang2026utonia}   & $\sim$250k + 1M CAD & 100$^\dagger$ / 256 / 64 H20 & yes & cross-dom. \\
    \method{} & $12.8$k (22 sets) & 50$^\ddagger$ / 512 / 128 & no & outdoor/city \\
    \bottomrule
  \end{tabular}\\[0.25em]
  {\scriptsize $^\dagger$Utonia Stage~2 (Stage~1 length unstated).
  $^\ddagger$Epoch-50 probe checkpoint, selected near the early validation
    plateau; some runs continued to 150 epochs to assess convergence.
    Global batch: Sonata 96, Utonia 256, \method{} $4{\times}128$ (=512).
    Concerto batch unstated.}
\end{table}

\clearpage
\section{Method}
\label{sec:method}

\subsection{Overview}

\method{} is implemented as four coupled elements: an outdoor corpus mixture, spatial and gravity-preserving view sampling, the LitePT-L backbone, and a single-encoder, distributionally-regularized objective
(\cref{fig:architecture}).
Our cross-model experiments evaluate this combination as a unit.
The encoder is trained to align global and local tokens that have been paired in the original, pre-sampling coordinate frame, while matching a regularized feature distribution:
{\setlength{\belowdisplayskip}{6pt plus 1pt minus 2pt}%
\begin{equation}
\begin{aligned}
\mathcal{L}
&= \lambda\,\mathcal{L}_{\mathrm{SIGReg}}
 + (1{-}\lambda)\,\mathcal{L}_{\mathrm{inv}} \\
&\quad + w_{\mathrm{f}}\mathcal{L}_{\mathrm{var\text{-}f}}
 + w_{\mathrm{c}}\mathcal{L}_{\mathrm{var\text{-}c}}
 + w_{\mathrm{cov}}\mathcal{L}_{\mathrm{cov}},
\end{aligned}
\label{eq:total-loss}
\end{equation}}
where the terms are defined in \cref{sec:loss} and numerical settings are collected in \cref{app:hparams}.
Adding VICReg-style terms gives up vanilla SIGReg's attractive single-hyperparameter formulation, but
we accept this trade-off because each term has an explicit role in controlling collapse, feature scale, or redundancy.
This remains easier to interpret than DINO-style objectives, whose loss values can be difficult to relate to downstream transfer and may not decrease consistently during training.

\vspace{-3pt}
\subsection{Training Corpus}
\label{sec:data}

Pretraining uses 22 outdoor and city-scale corpora under a unified loader pipeline, spanning aerial imagery, LiDAR, photogrammetry, meshes, heritage buildings (some with interiors), and driving sequences; the full dataset list and local-release provenance are provided in \cref{app:pretrain-corpus}.
The headline run uses $12{,}759$ scenes after capping excessive FRACTAL forest scenery and removing some broken scenes.
Sequential LiDAR datasets pool at most eight frames per access but remain in the scene tally, to avoid overpowering the city-scale scenes in the data mixture.
\Cref{tab:pretrain-scale} contrasts this outdoor pool with the published baseline corpora.
Inputs are the LitePT default 8-D feature vector of XYZ, RGB, normals, and a binary presence indicator for missing or unavailable modalities; because very few of the training datasets included normals, Polis was trained and runs entirely without them.

To keep large corpora from dominating, a corpus with $n_d$ indexed scenes is sampled with
\begin{equation}
  p(d)=\frac{n_d^{\alpha}}{\sum_k n_k^{\alpha}}, \qquad \alpha=0.15,
  \label{eq:corpus-sampling}
\end{equation}
after the stated corpus caps are applied.
When necessary, very large scenes are first partitioned into persistent spatial chunks that can be loaded independently.
From each scene or chunk, an area crop selects a nearest-neighbor region around a sampled center ($51{,}036$ area-crop samples per epoch; four crops per scene); the global and local views of \cref{sec:augmentations} are then sampled from that crop.

\subsection{Multi-View Augmentations}
\label{sec:augmentations}

We follow standard joint embedding practice by utilizing a multi-crop, multi-view schedule~\cite{caron2021dino}: one broad global view and two smaller local views are sampled from each area crop at each training pass. 
Each view receives independent geometric and chromatic perturbations, including differential yaw, tilt, scale, flips, jitter, elastic distortion~\cite{choy2019minkowski}, and cuboid dropout (a 3D analog of cutout / random erasing)~\cite{devries2017cutout,zhong2020random}.
On scene-scale data, rather than object-scale, Utonia likewise keeps gravity: full yaw $z\in[-\pi,\pi]$ with only mild roll/pitch $x/y\in[-\pi/64,\pi/64]$, and reserves full SO(3) rotations for object-centric samples~\cite{zhang2026utonia}.
\method{} never applies those object-level SO(3) rotations, because its corpus is outdoor and city-scale.
The remaining difference is view-asymmetric yaw: local views always receive a $z$ rotation, while global views do so with probability $0.5$, so the global view often retains the scene's original heading; $x/y$ tilts remain $\pm\pi/64$ ($\approx\pm 2.81^\circ$) on both views.
The exact view scales, point cap, and optimization settings are listed in \cref{app:hparams}.

The crop, view sampling, and augmentations above apply only during pretraining; probe-time decoding is defined in \cref{sec:experiments}.

\vspace{-1pt}
\subsection{Objective}
\label{sec:loss}
\begingroup
\setlength{\abovedisplayskip}{6pt plus 2pt minus 2pt}
\setlength{\belowdisplayskip}{6pt plus 2pt minus 2pt}
\setlength{\abovedisplayshortskip}{0pt plus 2pt}
\setlength{\belowdisplayshortskip}{4pt plus 2pt minus 1pt}
\setlength{\jot}{2pt}

Let $z_i^G,z_i^L\in\mathbb{R}^D$ be a valid global--local feature pair and let $\mathcal{M}$ contain all such pairs in the batch.
The invariance term is thus: 
\begin{equation}
  \mathcal{L}_{\mathrm{inv}}
  =\frac{1}{|\mathcal{M}|}\sum_{i\in\mathcal{M}}
  \left(1-\frac{\langle z_i^G,z_i^L\rangle}
  {\|z_i^G\|_2\|z_i^L\|_2}\right).
  \label{eq:invariance}
\end{equation}
Because the global and local views are sampled, voxelized, and augmented independently, their tokens do not have one-to-one index correspondence.
We retain each token's original, pre-augmentation coordinate and match every local token to its nearest global token in that coordinate frame.
Matches farther than three voxel widths ($0.06\,\mathrm{m}$) are discarded, so \cref{eq:invariance} aligns only spatially corresponding content visible in both views.
More than $80\%$ of local tokens satisfy this criterion for most scenes.
This geometric correspondence is a 3D-specific pairing rule; the cosine loss aligns the matched tokens directly, rather than through a latent predictor.

We apply SIGReg~\cite{balestriero2025lejepa} only to the final-stage tokens of the encoder, giving the earlier stages more representational flexibility while reducing computational load.
SIGReg is the average of a univariate Epps--Pulley statistic $T_{\mathrm{EP}}$ over $M$ random unit directions $a_m$:
\begin{equation}
  \mathcal{S}(Z)
  =\frac{1}{M}\sum_{m=1}^{M} T_{\mathrm{EP}}(Z a_m),
  \qquad M=256.
  \label{eq:sigreg-s}
\end{equation}
It is evaluated separately on the set of global-view tokens and the set formed by concatenating both local views, then averaged:
{\setlength{\abovedisplayskip}{3pt plus 1pt minus 1pt}%
\begin{equation}
  \mathcal{L}_{\mathrm{SIGReg}}
  =\frac{1}{2}\sum_{B\in\{G,\,L\}}\mathcal{S}(Z^{B}).
  \label{eq:sigreg}
\end{equation}}

For each token set $B\in\{G,L\}$, let $\sigma_j^B$ be the batch standard deviation of feature dimension $j$ and $C^B$ its feature correlation matrix.
The additional anti-collapse terms are:
\begin{align}
\mathcal{L}_{\mathrm{var\text{-}f}}
&=\frac{1}{2D}\sum_{B,j}[\gamma-\sigma_j^B]_+,\nonumber\\
\mathcal{L}_{\mathrm{var\text{-}c}}
&=\frac{1}{2D}\sum_{B,j}[\sigma_j^B-\gamma]_+,\nonumber\\
\mathcal{L}_{\mathrm{cov}}
&=\frac{1}{2D(D-1)}\sum_B\sum_{j\ne k}(C_{jk}^B)^2,
\label{eq:anticollapse}
\end{align}
where $[x]_+=\max(0,x)$.
Equal floor/ceiling targets $\gamma$ pull each dimension toward $\sigma_j^B=\gamma$; the off-diagonal penalty on $C^B$ penalizes redundancy.
We retain them because cosine invariance does not constrain scale, and because removing them hurts stability and transfer (\cref{sec:ablation-ema-pred}).
LeJEPA warns that pairing $\ell_2$ invariance with many SIGReg slices can recover a VICReg-like shortcut~\cite{balestriero2025lejepa}; we therefore use cosine matching and $M=256$, before testing for residual geometric shortcuts in \cref{sec:radial} to verify the representations did not collapse.

The locked fourteen-dataset evaluation used no test metric for selection; optimizer, precision, distributed-training, and exposure details are given in \cref{app:hparams}.
\endgroup

\vspace{-6pt}
\section{Experiments}
\label{sec:experiments}
\enlargethispage{1\baselineskip}

\subsection{Semantic Segmentation Probing Protocol}

We evaluate frozen-feature semantic segmentation on fourteen outdoor city- and building-scale corpora spanning the capture modalities and urban styles of real deployments; the full list is visible in \cref{tab:main-results} and \cref{fig:miou-transfer}.
We use the city- and building-scale split from \cref{sec:intro}; SemanticUrban~\cite{fang2026semanticurban} is building-scale, localized terrestrial LiDAR despite its urban subject matter.
SensatUrban~\cite{hu2022sensaturban} is validation-only because public test labels are unavailable.
As our probing head, we adapt Efficient Probing (EP)~\cite{efficientprobing} from global classification to dense segmentation: eight learned queries pool a global descriptor from frozen final-stage features, which is concatenated with each local feature and decoded by a two-layer MLP; only the EP pooler and MLP are trained.
Unless noted, probes run for 10 epochs, and we select the probe checkpoint applied to the test sets by validation Acc.\ when available.
Baselines use official Sonata, Concerto, and Utonia encoders under the same probe codepath but with native preprocessing ($2\,$cm voxels for Polis/Sonata/Concerto; $1\,$cm for Utonia).
Each probe pass is limited by a point cap (after deterministic subsampling) and a voxel cap (occupied voxels at the native grid); the two are not interchangeable.
Matched-budget probing uses a shared $400\mathrm{k}$-point / $262\mathrm{k}$-voxel limit with overlapping windows; high-capacity probing uses each model's largest practical pair with one-shot decoding of a globally subsampled scene.
The windowed protocol is slower than one-shot evaluation, because each scene is encoded in multiple spatial passes rather than a single forward pass.
A randomly initialized LitePT-L scores $5.7\%$ vs.\ $23.5\%$ mIoU on DALES ($3.7\%$ vs.\ $16.7\%$ over fourteen datasets) under the same probe, and fully supervised KPConv on DALES reaches $81.1\%$~\cite{thomas2019kpconv,varney2020dales}, so frozen scores reflect pretrained features rather than the probe, while remaining well below task-specific aerial segmentation.
We first ablate objective components on a smaller corpus (\cref{sec:ablation-ema-pred}), then report full-system transfer.

\subsection{Objective Ablations}
\label{sec:ablation-ema-pred}
\enlargethispage{2\baselineskip}

The LeJEPA motivation is that an EMA teacher and a latent predictor may be unnecessary once SIGReg constrains the embedding. We test this on a smaller outdoor corpus using the same LitePT-L backbone and fixed 50-epoch pretraining budget, repeated over three seeds ($0$, $7$, $40$). Every variant is then frozen and probed on the same six datasets with one-shot high-capacity decoding. These scores are comparable only within this smaller-corpus ablation, not to the full-corpus results in \cref{tab:main-results}.

\vspace{3pt}
\noindent
\begin{minipage}{\linewidth}
  \centering
  \captionsetup{skip=6pt}
  \captionof{table}{Small-corpus ablations over three seeds (mean$\pm$std test mIoU, \%).
  City-4: DALES, Swiss3DCities, Turin3D, YTU3D; All-6 adds ArCH and UnderOneFacade for building-scale evaluation.
  Dagger: losses matched to the headline \method{} recipe.
  The $\ell_2$ (MSE) invariance row retains SIGReg and the anti-collapse terms, changing only the matched-token loss.
  The EMA-teacher style uses those losses without temperature scaling or centering; Sonata-style replicates the full self-distillation training style, including the objectives.}
  \label{tab:small-ablation}
  \footnotesize
  \renewcommand{\arraystretch}{1.0}
  \begin{tabular*}{\linewidth}{@{\extracolsep{\fill}}lcc@{}}
    \toprule
    Variant & City-4 mIoU & All-6 mIoU \\
    \midrule
    Encoder-only$^\dagger$ & $\mathbf{21.01\pm1.55}$ & $\mathbf{14.86\pm1.11}$ \\
    Predictor$^\dagger$ & $19.53\pm0.60$ & $14.12\pm0.48$ \\
    No variance terms & $19.22\pm1.23$ & $14.84\pm0.79$ \\
    $\ell_2$ invariance & $18.95\pm1.17$ & $13.88\pm0.40$ \\
    No covariance terms & $17.14\pm0.29$ & $12.56\pm0.14$ \\
    No anti-collapse terms & $14.70\pm2.87$ & $10.42\pm2.12$ \\
    Sonata-style distillation & $14.11\pm0.71$ & $12.83\pm0.57$ \\
    EMA teacher$^\dagger$ & $4.47\pm1.73$ & $3.31\pm1.15$ \\
    \bottomrule
  \end{tabular*}
\end{minipage}
\vspace{1pt}

Encoder-only leads both City-4 and All-6, but the predictor is close enough that three seeds do not establish a meaningful accuracy gain from removing it; we keep the encoder-only design because it is simpler, not because predictor removal produces very strong outperformance.
The EMA-teacher uses loss terms matched to the encoder-only setup without temperature scaling or centering; its collapse therefore indicts this hybrid configuration rather than EMA teachers in general.
Removing all anti-collapse terms remains weak, while ablating covariance or variance alone stays intermediate.
Replacing cosine matching with $\ell_2$ (MSE) invariance, while retaining SIGReg and the anti-collapse terms, trails encoder-only on both aggregates.

\subsection{Main Results}
\label{sec:main-results}

\begin{figure*}[!t]
  \centering
  \includegraphics[width=\textwidth,height=0.28\textheight,keepaspectratio]{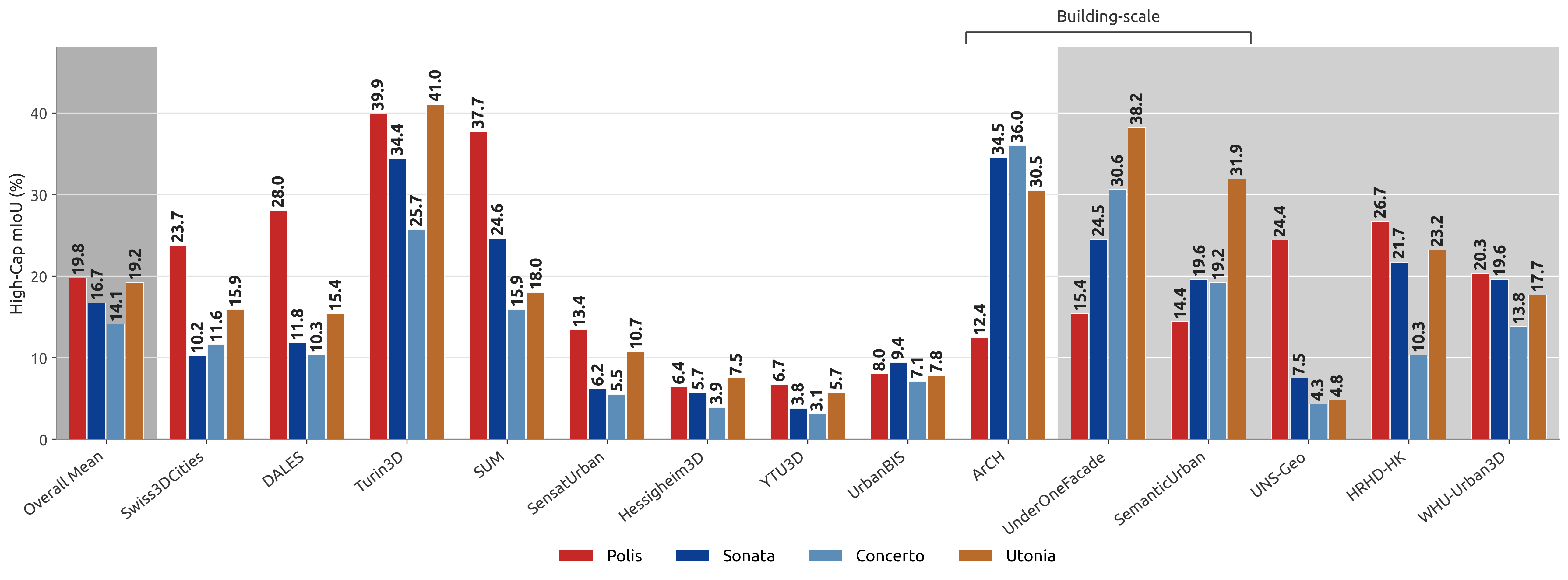}
  \caption{High-cap frozen-probe mIoU across 14 corpora (dark gray: cross-dataset mean; light gray: pretraining-disjoint held-out sets; top rule: building-scale corpora, the rest city-scale).
  Caps: Polis $1.5\mathrm{M}/1\mathrm{M}$, Sonata $1.0\mathrm{M}/655\mathrm{k}$, Concerto $600\mathrm{k}/393\mathrm{k}$, Utonia $800\mathrm{k}/524\mathrm{k}$.
  \textbf{Polis leads on broad city-scale scenes; Utonia retains the building-scale edge.}}
  \label{fig:miou-transfer}
\end{figure*}

\begin{table}[!b]
  \centering
  \caption{Matched-budget, native-preprocessing mIoU (\%) at 400k points / 262k voxels (10 probe epochs, best validation Acc. when available, otherwise best training Acc.).
  A superscript $^\dagger$ marks a pretraining-disjoint held-out dataset; SensatUrban is validation-only because public test labels are unavailable.}
  \label{tab:main-results}
  \scriptsize
  \renewcommand{\arraystretch}{0.9}
  \begin{tabular*}{\linewidth}{@{\extracolsep{\fill}}lcccc@{}}
    \toprule
    Dataset & \method{} & Sonata & Concerto & Utonia \\
    \midrule
    \multicolumn{5}{@{}l}{\emph{City-scale}} \\
    \addlinespace[0.15em]
    DALES~\cite{varney2020dales} & \textbf{23.5} & 9.0  & 8.2  & \underline{15.3} \\
    Swiss3DCities~\cite{can2021swiss3dcities} & \textbf{20.7} & 13.4 & \underline{17.4} & 14.7 \\
    Turin3D~\cite{barco2025turin3d} & \textbf{32.8} & 25.1 & 23.3 & \underline{29.2} \\
    SUM~\cite{gao2021sum} & \textbf{28.2} & \underline{20.8} & 11.3 & 16.3 \\
    YTU3D~\cite{bayrak2023ytu3d} & \textbf{5.6}  & 3.0  & 2.7  & \underline{5.2}  \\
    Hessigheim3D~\cite{kolle2021hessigheim} & \textbf{5.8}  & 3.9  & 3.5  & \underline{5.0}  \\
    UrbanBIS~\cite{yang2023urbanbis} & \underline{7.2} & \textbf{8.5} & 5.0 & 7.9 \\
    SensatUrban~\cite{hu2022sensaturban} & \textbf{12.4} & 6.6  & 6.0  & \underline{7.6} \\
    UNS-Geo$^\dagger$~\cite{govedarica2025unsgeo} & \underline{7.7} & 4.0  & 3.3  & \textbf{9.2} \\
    HRHD-HK$^\dagger$~\cite{li2023hrhd} & \textbf{24.1} & 11.5 & 17.5 & \underline{23.9} \\
    WHU-Urban3D$^\dagger$~\cite{han2024whuurban3d} & \textbf{20.1} & \underline{17.5} & 11.7 & 15.1 \\
    \midrule
    City held-out mean & \textbf{17.3} & 11.0 & 10.8 & \underline{16.1} \\
    All city mean   & \textbf{17.1} & 11.2 & 10.0 & \underline{13.6} \\
    \midrule
    \multicolumn{5}{@{}l}{\emph{Building-scale}} \\
    \addlinespace[0.15em]
    ArCH~\cite{matrone2020arch} & 15.1          & \textbf{34.1} & \underline{32.6} & 25.2 \\
    SemanticUrban$^\dagger$~\cite{fang2026semanticurban} & 13.6 & 16.9 & \underline{17.3} & \textbf{25.6} \\
    UnderOneFacade$^\dagger$~\cite{wang2026underonefacade} & 16.5 & 20.6 & \underline{23.8} & \textbf{29.3} \\
    \midrule
    Building held-out mean & 15.1      & 18.8 & \underline{20.6} & \textbf{27.5} \\
    All building mean & 15.1        & 23.9 & \underline{24.6} & \textbf{26.7} \\
    \midrule
    Overall mean    & \textbf{16.7} & 13.9 & 13.1 & \underline{16.4} \\
    \bottomrule
  \end{tabular*}
\end{table}

\Cref{fig:miou-transfer} summarizes the high-capacity probing mIoU across eleven city-scale datasets and three building-scale datasets; \cref{tab:main-results} reports the matched-budget values for the expanded comparison.
We test all models on nine datasets whose training sets \method{} saw during pretraining as well as five held-out datasets, which are delineated by $^\dagger$ marks in \cref{tab:main-results}.

The two protocols reveal the same city-scale specialization at different strengths.
Under high-capacity decoding, \method{} reaches $21.4\%$ mean mIoU across the eleven city-scale datasets, compared with $15.2\%$ for the next-best model, Utonia.
Its overall per-dataset lead is much narrower ($19.8\%$ versus $19.2\%$), reflecting its weaker transfer to localized terrestrial datasets.
The matched-budget comparison preserves this pattern: \method{} and Utonia are nearly tied overall ($16.7\%$ versus $16.4\%$), but \method{} retains a clearer city-scale lead ($17.1\%$ versus $13.6\%$).
The gain therefore does not arise solely from \method{}'s larger practical encoding capacity. \method{} leads on the
pretraining-disjoint broad-extent city datasets, while, conversely, the localized terrestrial sets favor the broader baselines. ArCH and UnderOneFacade label architectural elements such as walls, windows, and doors;
SemanticUrban uses a fine-grained streetscape taxonomy over single-station
captures of streets, squares, trees, and facades. Sonata, Concerto, and Utonia
therefore retain an advantage where small spatial extent, terrestrial viewpoint,
and part-level discrimination coincide.

\begin{figure*}[!t]
  \centering
  \includegraphics[width=\textwidth]{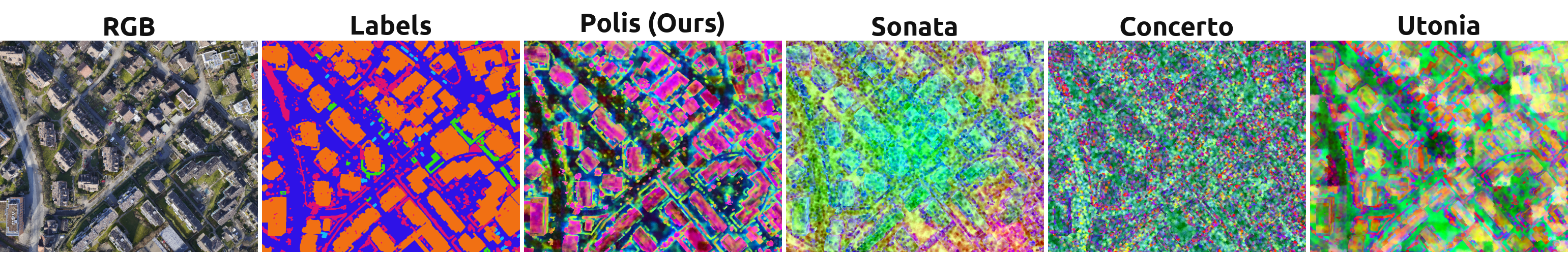}
  \caption{Top-down frozen-backbone feature PCA on a Zug scene from Swiss3DCities under the same max-fill qualitative encode protocol as \cref{fig:qual-zurich} using $k{=}32$ feature smoothing and 1-NN color fill for unsampled points.
  Labels use saturated categorical colors as a per-scene key, not a cross-model encoding.
  Sonata shows center color bias; Utonia shows axis-aligned patches.
  \textbf{These visual artifacts, along with the clear height direction artifacts present in \cref{fig:qual-zurich} for Utonia and Polis, motivate testing spatial layout and gravity-related cues as shortcuts in \cref{sec:radial}.}}
  \label{fig:qual-zug}
\end{figure*}

\begin{figure*}[!t]
  \centering
  \includegraphics[width=0.92\textwidth]{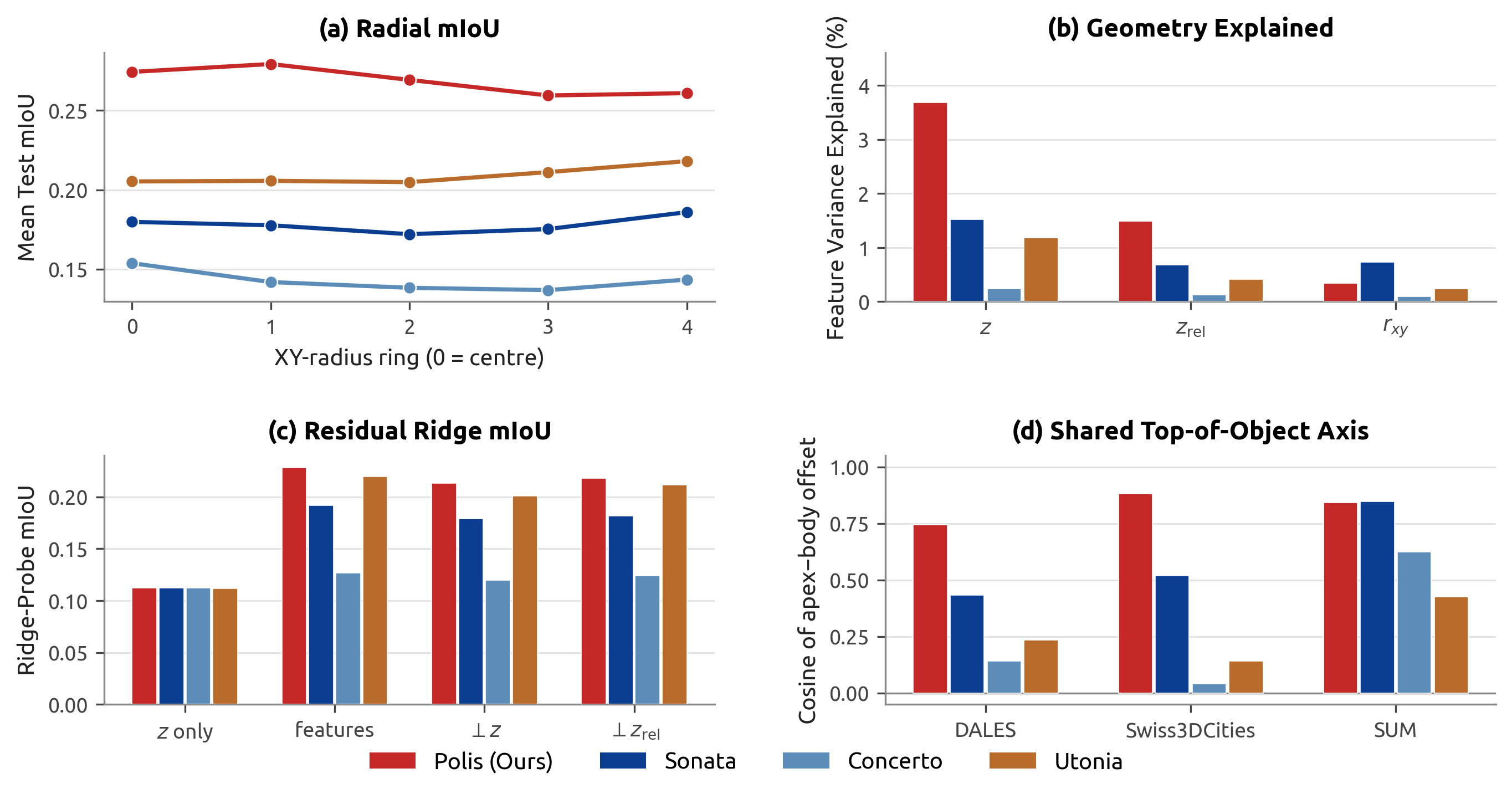}
  \caption{Radial mIoU and geometric structure of frozen features.
  (a)~Locked high-capacity one-shot test mIoU vs.\ XY-radius ring (macro-averaged over DALES, Swiss3DCities, SUM).
  (b)~Feature variance explained by $z$, $z_{\mathrm{rel}}$, and $r_{xy}$.
  (c)~Ridge mIoU before/after residualizing height cues.
  (d)~Shared building/vegetation apex-minus-body direction.
  \textbf{Polis encodes stronger gravity-aligned structure, but its city advantage does not reduce to these measured directions.}}
  \label{fig:radial}
\end{figure*}

\subsection{Qualitative Spatial Artifacts}
\label{sec:qual-artifacts}

The Zurich PCA qualitative render (\cref{fig:qual-zurich}) shows more coherent separation of urban objects and ground for Polis, including a shared top-of-object direction.
In the top-down view of Zug (\cref{fig:qual-zug}), Sonata shows a center color bias, while Utonia shows axis-aligned patches.
Following Sonata~\cite{wu2025sonata}, we call such reliance on accessible spatial cues a \emph{geometric shortcut} and test it empirically in \cref{sec:radial}.

\subsection{Radial Probe Accuracy and Geometric Shortcuts}
\label{sec:radial}

Under high-capacity one-shot decoding of large tiles, we bin test points into five equal-count XY-radius rings and macro-average locked-probe mIoU over DALES, Swiss3DCities, and SUM (\cref{fig:radial}a).
\method{} remains above the baselines across rings.
In that three-dataset mean, Sonata and Utonia are slightly weaker at the tile center than at the margin, whereas \method{} is slightly stronger at the center.
That center accuracy does not appear as a radial color field in the qualitative PCA (\cref{fig:qual-zug}): the visible center-color artifact is Sonata's, consistent with Sonata having the largest fraction of feature variance linearly explained by $r_{xy}$ ($0.73\%$ vs.\ $0.33\%$/$0.24\%$/$0.09\%$ for \method{}/Utonia/Concerto; \cref{fig:radial}b).
Lower $r_{xy}$ $R^2$ therefore argues against a linear XY shortcut in \method{}, but it does not by itself produce the center mIoU gain: Utonia encodes even less $r_{xy}$ and remains weaker at the center.
A more likely contributor is label layout in the tiles: Swiss3DCities favors the center for every backbone, while results on SUM are stronger at the margin, so a model that reads semantics rather than radius can still inherit a center or margin advantage from where the classes sit.
Class-conditional spherical variance is nearly flat across rings (see \cref{app:ring-var}), so the remaining Swiss drop is not explained by feature wash-out at the margins.

When understanding buildings and broad aerial city scenes, height is itself semantically meaningful.
Absolute height explains $3.69\%$ of \method{} feature variance versus $1.52\%$/$1.18\%$/$0.24\%$ for Sonata/Utonia/Concerto (\cref{fig:radial}b).
This stronger height encoding is consistent with training only on gravity-aligned outdoor scenes and never applying Utonia's object-level SO(3) rotations, which Utonia uses to weaken $z$-axis correlation while still preserving upright structure on scene-scale data.
To test whether that encoding is a reusable top-of-object axis rather than a preference for globally high points, we isolate a relative-height coordinate $z_{\mathrm{rel}}$.
Within each class-consistent connected component with at least $3\mathrm{m}$ of vertical span, $z_{\mathrm{rel}}$ is the point's height normalized by that component's own min--max, so a house roof and a skyscraper roof can both sit near $z_{\mathrm{rel}}=1$.
We then take, for buildings and for vegetation, the mean frozen feature of apex points ($z_{\mathrm{rel}}\ge 0.90$) minus the mean feature of body points ($0.20\le z_{\mathrm{rel}}\le 0.70$), and report the cosine of those two offset vectors (\cref{fig:radial}d).
A high value means ``moving toward the top of an object'' is a shared direction across classes; it does not mean that apex and body occupy unrelated regions of feature space.
\method{} has the most consistent shared offset on DALES and Swiss3DCities.
That axis is also visible in the Zurich PCA (\cref{fig:qual-zurich}): the roof of the top-left skyscraper shares the strong red of other rooftops, despite being much taller in absolute height, while the tops of some large trees likewise display a pink or red hue at their apex.
The geometric residualization ridge probe on held-out scenes (\cref{fig:radial}c) is a diagnostic separate from the locked test rings in \cref{fig:radial}a, so backbone order or absolute scores need not agree.
Even so, \method{} remains highest after removing the linear $\{z,z^2\}$ or $z_{\mathrm{rel}}$ components ($0.213$/$0.218$ vs.\ Utonia $0.201$/$0.212$), so the city advantage does not reduce to these measured vertical directions.

\subsection{Density Stability and Encode Max-Fill}
\label{sec:sparsity}

\begin{figure}[t]
  \centering
  \includegraphics[width=\linewidth]{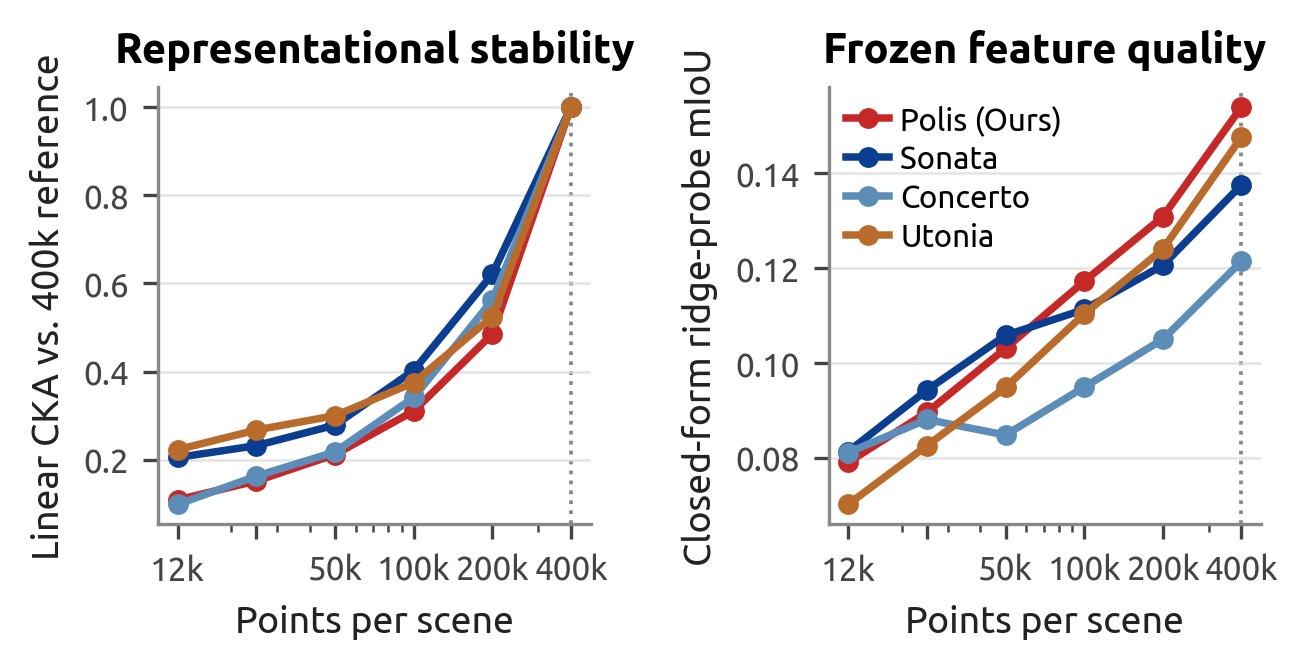}
  \caption{Frozen-feature density sweep on four scenes from each of our corpora.
  Left: linear CKA vs.\ the $400\mathrm{k}$-point encoding.
  Right: closed-form ridge probe mIoU.
  \method{} moves more under decimation than Sonata / Concerto / Utonia while remaining competitive on ridge mIoU.
  \textbf{Polis is more sensitive to decimation, yet demonstrates competitive transfer at practical point budgets.}}
  \label{fig:sparsity}
\end{figure}

To ask how the sparsity associated with downsampling modifies frozen features, we encode nested whole-scene random subsamples from $12.5\mathrm{k}$ to $400\mathrm{k}$ points (unbounded voxel cap), compute linear CKA against the $400\mathrm{k}$ encoding on shared points, and fit a closed-form ridge probe on the same features (\cref{fig:sparsity}).
Feature stability and downstream transfer follow the same density-dependent pattern.
\method{} has the lowest CKA similarity to its $400\mathrm{k}$-point, matched-budget reference under decimation~\cite{kornblith2019cka} and trails one or more baselines at several low-density ridge probes.
This sensitivity is potentially an artifact of limited sparsity augmentations in \method{}'s pretraining recipe paired with a higher variance across the sparsity of the input datasets, allowing the model to learn a density shortcut for identifying particular datasets. Despite this decimation sensitivity, \method{} remains competitive on ridge mIoU, even at the lowest densities, showing that the lack of between-density consistency in the feature space does not mean the destruction of semantics.
Budgets above $400\mathrm{k}$ are omitted from the curve because several tiles (notably Hessigheim3D) cannot host that many points with unbounded voxel capacity, so denser ladders drop corpora unevenly across backbones.
Under one shared unmodified \texttt{spconv}/\texttt{cumm} runtime on an H100 80\,GB GPU (\cref{fig:capacity-transfer}), median one-shot maxima are $1.5\mathrm{M}$ points for \method{}, $1.0\mathrm{M}$ for Sonata/Utonia, and ${\sim}0.76\mathrm{M}$ for Concerto (a 32-bit kernel index ceiling, not device memory; see \cref{app:spconv}). We keep this unpatched runtime because it is the official wheel that ships with Sonata, Concerto, and Utonia.

\section{Discussion and Limitations}
\label{sec:limitations}

The outdoor-specialized system transfers best on broad aerial city scenes and is weaker on localized terrestrial streetscape and single-building scenes.
Capture geometry, scene extent, and taxonomy granularity covary in this suite, so their individual causal contributions remain unresolved.
SIGReg makes the single-encoder recipe viable, but the city lead is a property of the complete system; we keep the encoder-only design instead of using a predictor because it is simpler, not because the small-corpus ablation isolates it as the source of better features.
\vspace{-3pt}
\section{Conclusion}
\label{sec:conclusion} 

City-scale 3D understanding needs more than an indoor or object-trained encoder applied to a larger scene.
\method{} is designed around that premise: an encoder-only, distributionally-regularized
objective with geometrically matched views, outdoor data, and gravity-preserving
augmentation. This yields $23.8\%$ mean mIoU on three pretraining-disjoint city
datasets versus $16.3\%$ for the next-best encoder under high-capacity probing,
and $17.3\%$ versus $16.1\%$ at a matched point and voxel budget with spatially windowed decoding.

This makes \method{} useful for city-scale applications in which the scene
itself is part of the signal, rather than merely a container for isolated
objects. The trade-off is equally important: on localized terrestrial captures
with fine-grained building-part and streetscape labels, \method{} reaches
$14.1\%$ mIoU versus $33.5\%$ for the strongest baseline. Polis therefore
advances outdoor 3D self-supervision not by replacing general-purpose
representations, but by showing how a representation can be deliberately
aligned with the geometry and context of a target deployment regime. The next
step is to retain this city-scale advantage across the semantic hierarchy, from
broad urban structure to fine-grained building-part and streetscape labels, so
outdoor specialization does not come at the expense of the terrestrial detail
that broader pretraining provides.

\FloatBarrier

{
    \small
    \bibliographystyle{ieeenat_nonumber}
    \bibliography{references}
}

\clearpage
\appendix
\input{appendix}

\end{document}

%% file: appendix.tex

\section{Pretraining Corpus}
\label[appendix]{app:pretrain-corpus}

Pretraining uses 22 outdoor and city-scale corpora under a unified loader.
The indexed sources are Aerial-MegaDepth~\cite{vuong2025aerialmegadepth},
ArCH~\cite{matrone2020arch}, DALES~\cite{varney2020dales},
Dortmund3D~\cite{nex2015isprsbenchmark}, ECLAIR~\cite{melekhov2024eclair},
FRACTAL~\cite{gaydon2024fractal}, GauUScene~\cite{xiong2024gauuscene},
Hessigheim3D~\cite{kolle2021hessigheim}, MatrixCity~\cite{li2023matrixcity},
STPLS3D~\cite{chen2022stpls3d}, SUM~\cite{gao2021sum},
SemanticKITTI~\cite{behley2019semantickitti},
SensatUrban~\cite{hu2022sensaturban}, Swiss3DCities~\cite{can2021swiss3dcities},
TUM-Facade~\cite{wysocki2022tumfacade}, TerraSky3D~\cite{durso2026terrasky3d},
Turin3D~\cite{barco2025turin3d}, UrbanBIS~\cite{yang2023urbanbis},
UseGeo~\cite{nex2024usegeo}, VBR~\cite{brizi2024vbr},
Venice~\cite{dilenardo2015venicetime,dilenardo2025sanmarco4d,guhennec2023skinvenice},
and YTU3D~\cite{bayrak2023ytu3d}.
Mix3D scene mixing~\cite{nekrasov2021mix3d} is implemented in the loader but
was disabled for this run.

The headline run indexes $12{,}759$ on-disk scenes after caps and blocklisting
($51{,}036$ area-crop samples per epoch; four crops per scene).
The FRACTAL hard cap reduces $90{,}799$ scenes to $12{,}799$, keeping
$2{,}000$ FRACTAL scenes; a subsequent $40$-scene blocklist of broken
Venice/SensatUrban files gives the final $12{,}759$.
SemanticKITTI and VBR pool at most eight frames per access but remain in the
scene tally.
Dortmund3D is the photogrammetric point cloud from the ISPRS/EuroSDR
Benchmark for Multi-Platform Photogrammetry (Dortmund city-centre and Zeche
Zollern)~\cite{nex2015isprsbenchmark}, released at
\url{https://www.isprs.org/resources/datasets/benchmarks/MultiPlatformPhoto/Default.aspx}.
Venice follows the Venice Time Machine / TimeAtlas point-cloud
lineage~\cite{dilenardo2015venicetime,dilenardo2025sanmarco4d,guhennec2023skinvenice}
and is not a public benchmark release.

\section{Supervised Calibration}
\label[appendix]{app:supervised}

For reference, the DALES paper reports fully supervised
KPConv~\cite{thomas2019kpconv,varney2020dales} at $81.1\%$ mIoU. This end-to-end, task-specific result is not directly comparable
to our frozen, lightweight probes: those use a small head trained for 10 epochs
on at most 1000 scenes, with substantial downsampling on tiles that often exceed
$10^7$ points. We report the supervised number only as a rough scale check for
the absolute frozen-probe scores under the constrained probe protocol.

\section{Evaluation Corpora}
\label[appendix]{app:eval-corpora}

City-scale probe sets with public labels:
DALES~\cite{varney2020dales}, Swiss3DCities~\cite{can2021swiss3dcities},
Turin3D~\cite{barco2025turin3d}, SUM~\cite{gao2021sum},
YTU3D~\cite{bayrak2023ytu3d}, Hessigheim3D~\cite{kolle2021hessigheim},
UrbanBIS~\cite{yang2023urbanbis}, UNS-Geo~\cite{govedarica2025unsgeo},
HRHD-HK~\cite{li2023hrhd}, and WHU-Urban3D~\cite{han2024whuurban3d}.
Building-scale sets: ArCH~\cite{matrone2020arch},
UnderOneFacade~\cite{wang2026underonefacade}, and
SemanticUrban~\cite{fang2026semanticurban}.
SensatUrban~\cite{hu2022sensaturban} is validation-only because official test
labels are not public.

\section{Random-Initialization Probe}
\label[appendix]{app:random-init}

A randomly initialized LitePT-L under the same matched-budget frozen probe
as the main paper ($400\mathrm{k}$ points / $262\mathrm{k}$ voxels, 10
epochs, at most 1000 training scenes) is the calibration for the
fourteen-dataset overall mean cited in the main text ($3.7\%$ vs.\
$16.7\%$).
\Cref{tab:random-init} reports per-dataset mIoU.
Random sparse-convolution features are not a trivial baseline: they still
encode local voxel geometry, so the gap is the contribution of pretraining
rather than of the probe head.

\begin{table}[h]
  \centering
  \caption{Matched-budget frozen-probe mIoU (\%). Random-init is an
  untrained LitePT-L; Polis is the headline pretrained checkpoint.
  A superscript $^\dagger$ marks a pretraining-disjoint held-out dataset;
  SensatUrban is validation mIoU because public test labels are unavailable.}
  \label{tab:random-init}
  \footnotesize
  \begin{tabular}{@{}lcc@{}}
    \toprule
    Dataset & Random-init & Polis \\
    \midrule
    \multicolumn{3}{@{}l}{\emph{City-scale}} \\
    DALES~\cite{varney2020dales} & 5.7 & 23.5 \\
    Swiss3DCities~\cite{can2021swiss3dcities} & 6.8 & 20.7 \\
    Turin3D~\cite{barco2025turin3d} & 4.2 & 32.8 \\
    SUM~\cite{gao2021sum} & 7.0 & 28.2 \\
    YTU3D~\cite{bayrak2023ytu3d} & 0.9 & 5.6 \\
    Hessigheim3D~\cite{kolle2021hessigheim} & 3.3 & 5.8 \\
    UrbanBIS~\cite{yang2023urbanbis} & 2.1 & 7.2 \\
    SensatUrban~\cite{hu2022sensaturban} & 3.0 & 12.4 \\
    UNS-Geo$^\dagger$~\cite{govedarica2025unsgeo} & 3.3 & 7.7 \\
    HRHD-HK$^\dagger$~\cite{li2023hrhd} & 6.9 & 24.1 \\
    WHU-Urban3D$^\dagger$~\cite{han2024whuurban3d} & 4.7 & 20.1 \\
    \midrule
    All city mean & 4.4 & 17.1 \\
    \midrule
    \multicolumn{3}{@{}l}{\emph{Building-scale}} \\
    ArCH~\cite{matrone2020arch} & 0.0 & 15.1 \\
    SemanticUrban$^\dagger$~\cite{fang2026semanticurban} & 2.2 & 13.6 \\
    UnderOneFacade$^\dagger$~\cite{wang2026underonefacade} & 1.9 & 16.5 \\
    \midrule
    All building mean & 1.4 & 15.1 \\
    \midrule
    Overall mean & 3.7 & 16.7 \\
    \bottomrule
  \end{tabular}
\end{table}

\section{High-Capacity Probe}
\label[appendix]{app:high-cap}

\Cref{tab:high-cap} reports the high-capacity one-shot scores plotted in
\cref{fig:miou-transfer}, using the same dataset order, held-out marks,
and best~/~second-best highlighting as \cref{tab:main-results}.
Caps are model-specific, matching that figure: \method{} $1.5\mathrm{M}/1\mathrm{M}$,
Sonata $1.0\mathrm{M}/655\mathrm{k}$, Concerto $600\mathrm{k}/393\mathrm{k}$,
Utonia $800\mathrm{k}/524\mathrm{k}$.

\begin{table}[h]
  \centering
  \caption{High-capacity, native-preprocessing mIoU (\%) with model-specific one-shot caps (10 probe epochs, best validation Acc.\ when available, otherwise best training Acc.).
  A superscript $^\dagger$ marks a pretraining-disjoint held-out dataset; SensatUrban is validation-only because public test labels are unavailable.}
  \label{tab:high-cap}
  \scriptsize
  \renewcommand{\arraystretch}{0.9}
  \begin{tabular*}{\linewidth}{@{\extracolsep{\fill}}lcccc@{}}
    \toprule
    Dataset & \method{} & Sonata & Concerto & Utonia \\
    \midrule
    \multicolumn{5}{@{}l}{\emph{City-scale}} \\
    \addlinespace[0.15em]
    DALES~\cite{varney2020dales} & \textbf{28.0} & 11.8 & 10.3 & \underline{15.4} \\
    Swiss3DCities~\cite{can2021swiss3dcities} & \textbf{23.7} & 10.2 & 11.6 & \underline{15.9} \\
    Turin3D~\cite{barco2025turin3d} & \underline{39.9} & 34.4 & 25.7 & \textbf{41.0} \\
    SUM~\cite{gao2021sum} & \textbf{37.7} & \underline{24.6} & 15.9 & 18.0 \\
    YTU3D~\cite{bayrak2023ytu3d} & \textbf{6.7}  & 3.8  & 3.1  & \underline{5.7}  \\
    Hessigheim3D~\cite{kolle2021hessigheim} & \underline{6.4}  & 5.7  & 3.9  & \textbf{7.5}  \\
    UrbanBIS~\cite{yang2023urbanbis} & \underline{8.0} & \textbf{9.4} & 7.1 & 7.8 \\
    SensatUrban~\cite{hu2022sensaturban} & \textbf{13.4} & 6.2  & 5.5  & \underline{10.7} \\
    UNS-Geo$^\dagger$~\cite{govedarica2025unsgeo} & \textbf{24.4} & \underline{7.5} & 4.3 & 4.8 \\
    HRHD-HK$^\dagger$~\cite{li2023hrhd} & \textbf{26.7} & 21.7 & 10.3 & \underline{23.2} \\
    WHU-Urban3D$^\dagger$~\cite{han2024whuurban3d} & \textbf{20.3} & \underline{19.6} & 13.8 & 17.7 \\
    \midrule
    City held-out mean & \textbf{23.8} & \underline{16.3} & 9.5 & 15.2 \\
    All city mean   & \textbf{21.4} & 14.1 & 10.1 & \underline{15.2} \\
    \midrule
    \multicolumn{5}{@{}l}{\emph{Building-scale}} \\
    \addlinespace[0.15em]
    ArCH~\cite{matrone2020arch} & 12.4          & \underline{34.5} & \textbf{36.0} & 30.5 \\
    SemanticUrban$^\dagger$~\cite{fang2026semanticurban} & 14.4 & \underline{19.6} & 19.2 & \textbf{31.9} \\
    UnderOneFacade$^\dagger$~\cite{wang2026underonefacade} & 15.4 & 24.5 & \underline{30.6} & \textbf{38.2} \\
    \midrule
    Building held-out mean & 14.9      & 22.1 & \underline{24.9} & \textbf{35.0} \\
    All building mean & 14.1        & 26.2 & \underline{28.6} & \textbf{33.5} \\
    \midrule
    Overall mean    & \textbf{19.8} & 16.7 & 14.1 & \underline{19.2} \\
    \bottomrule
  \end{tabular*}
\end{table}

\section{Class-Conditional Spherical Variance}
\label[appendix]{app:ring-var}

The omitted panel from \cref{fig:radial} reports within-class
spherical variance of $\ell_2$-normalized frozen features, computed on the
high-capacity cached encodings and macro-averaged over classes with support
in each equal-count XY-radius ring.
Across DALES, Swiss3DCities, and SUM the ring curves are nearly flat: the
largest centre-to-margin range is $0.011$ (Utonia on SUM; Polis $0.010$ on
DALES), and Swiss3DCities ranges are at most $0.007$.
A Swiss centre-to-margin mIoU drop is therefore not explained by features
washing out at the tile margin.

\section{Hyperparameters}
\label[appendix]{app:hparams}

\begin{table}[h]
  \centering
  \caption{Hyperparameters for \method{} pretraining and frozen probing. The probe checkpoint, data sampling, objective weights, and resource settings are reported to make the training and evaluation protocol reproducible.}
  \footnotesize
  \begin{tabular}{@{}ll@{}}
    \toprule
    Parameter & Value \\
    \midrule
    Grid size & $0.02\,\mathrm{m}$ \\
    Views & 1 global + 2 local \\
    Global / local scales & $(0.4,1.0)$ / $(0.02,0.25)$ \\
    Points per view & $100\mathrm{k}$ (headline run) \\
    $\lambda$ (SIGReg) & $5{\times}10^{-4}$ \\
    $w_{\mathrm{f}} / w_{\mathrm{c}}$ (variance weights) & $0.3$ / $0.3$ \\
    $\gamma$ (std target) & $1.0$ \\
    $w_{\mathrm{cov}}$ & $3.5$ \\
    SIGReg slices / token cap & $256$ / $16\mathrm{k}$ \\
    Optimizer & AdamW, lr $10^{-4}$, wd $0.05$ \\
    Pretrain epochs & 150 (probe ckpt: 50) \\
    DDP & 128 GPUs, batch 4/GPU (global 512) \\
    Precision & bf16 \\
    Probe & 10 ep., best val Acc., 400k / 262k \\
    \bottomrule
  \end{tabular}
\end{table}

The recipe combines persistent spatial chunks, per-pass area crops, DINO global and local views, differential yaw, and batch-level rather than per-scene statistics.

\section{Sparse-Convolution Capacity Runtime}
\label[appendix]{app:spconv}

All four encoders were evaluated in the same environment using the official
PyPI wheels \texttt{spconv-cu124}~2.3.8 and
\texttt{cumm-cu124}~0.7.11.  This follows the released Sonata, Concerto, and
Utonia installation instructions, which request
\texttt{spconv-cu\$\{CUDA\_VERSION\}} (and use
\texttt{spconv-cu124} in their CUDA~12.4 environment files) without requiring
a custom branch.  At sufficiently large sparse tensors, the released
implicit-GEMM path rejects a temporary whose
$N{\times}C{\times}\text{dtype-bytes}$ extent exceeds a 32-bit byte-addressing
guard.  The resulting assertion is documented upstream in
\url{https://github.com/traveller59/spconv/issues/706}, with related reports in
\url{https://github.com/traveller59/spconv/issues/604} and
\url{https://github.com/traveller59/spconv/issues/656}.  This guard is distinct
from the library's existing 64-bit hashing support for large spatial shapes
and can therefore trigger while substantial GPU memory remains available.

Community remedies include rebuilding
\url{https://github.com/FindDefinition/cumm} with the generated NVRTC
\texttt{int\_max} and associated convolution assertions widened to 64 bits,
then rebuilding \url{https://github.com/traveller59/spconv}; the upstream
discussion demonstrates this using cumm~0.7.11 and spconv~2.3.8.  Development
forks are also available at
\url{https://github.com/chensjtu/cumm-int32} and
\url{https://github.com/chensjtu/spconv-int32}, but the latter explicitly
describes its workaround as inference-oriented and unsuitable for training.
Because these remedies are not official CUDA~12.4 wheels and would introduce
an additional systems variable, none was used for any model in our comparison.